\documentclass[12pt]{article}

\usepackage[a4paper,margin=1in]{geometry}
\usepackage{setspace}
\usepackage{graphicx}
\usepackage{amsmath}
\usepackage{amssymb}
\usepackage{url}
\usepackage{placeins}
\usepackage{tabularx}
\usepackage{comment}
\usepackage{xcolor}
\usepackage{hyperref}

\begin{document}

\title{Linked Data Classification using Neurochaos Learning}

\author{Pooja Honna, Ayush Patravali, Nithin Nagaraj,\\Nanjangud C.~Narendra\\Complex Systems Programme \\National Institute of Advanced Studies (NIAS) \\
IISc Campus, Bangalore 560 012, India\\
\textit{poojahonna24@gmail.com, ayushpatravali10@gmail.com}\\\textit{nithin@nias.res.in, ncnaren@gmail.com}}

\maketitle

\section{Abstract}

Neurochaos Learning (NL) has shown promise in recent times over traditional deep learning due to its two key features: ability to learn from small sized training samples, and low compute requirements. In prior work, NL has been implemented and extensively tested on separable and time series data, and  demonstrated its superior performance on both classification and regression tasks. In this paper, we investigate the next step in NL, viz., 
applying NL to \emph{linked data}, in particular, data that is represented in the form of knowledge graphs. We integrate linked data into NL by implementing node aggregation on knowledge graphs, and then feeding the aggregated node features to the simplest NL architecture: {\it ChaosNet}. We demonstrate the results of our implementation on homophilic graph datasets as well as heterophilic graph datasets of verying heterophily. We show better efficacy of our approach on homophilic graphs than on heterophilic graphs. While doing so, we also present our analysis of the results, as well as suggestions for future work.

\section{Introduction}\label{sec:intro}

Traditionally, deep learning~\cite{lecun2015deep} has relied on the neural network model. It has proved to be extremely successful for object recognition, speech recognition, classification and prediction, and has been used in several domains ranging from medicine to robotics~\cite{dong2021survey}.

However, deep learning has a few drawbacks~\cite{marcus2022deep}, such as correlation instead of causation, heavy dependence on data distribution and high energy consumption. To address these issues, we investigated the application of chaos theory to machine learning and utilized a recently proposed approach named Neurochaos Learning (NL)~\cite{harikrishnan2020neurochaos}. NL is based on the chaotic firing exhibited by neurons in the brain, and has proven to be effective at classification tasks, even with low training samples. Moreover, NL has also shown to require much lower compute than deep learning, rendering it usable even on regular CPUs instead of high-end energy-guzzling GPUs.

So far in the literature, research on NL has been limited to separable data. In this paper, we extend apply it to \emph{linked data}, i.e., data that is typically represented as a knowledge graph (KG). Linked data is more complex in that its analysis involves understanding and representing the \emph{relations} among the data points in the KG. This has usually been done using graph neural networks (GNN)~\cite{graph-class}. The first step in using GNNs has been to implement an aggregation of the features of the nodes of the graph, after which further processing is implemented on the aggregated data. In this paper, we adopt this approach by first aggregating the nodes in the KG, and then feeding the aggregated data as features to our NL system named ChaosNet. 

The testing of our approach was implemented on both homophilic (all nodes belong to the same class) and heterophilic (nodes belong to different classes) datasets. We obtained better results for homophilic datasets, and our results for heterophilic datasets degraded with increased heterophily of the dataset. 

The rest of this paper is organized as follows. We present some background information in the next Section. In Section~\ref{sec:litreview} we present and discuss some related work in this area. In Section~\ref{sec:researchgaps} we present the key research gaps that need to be addressed, some of which we will address in this paper. Section~\ref{sec:objectives} presents the goals and objectives against which we will measure our experimental results. Our experimental methodology is introduced in Section~\ref{sec:methodology}, and Section~\ref{sec:expts} will describe the experimental setup that was used. In Section~\ref{sec:results} we will present the results of our experiments on homophilic and heterophilic datasets; and we will analyze and discuss them in Section~\ref{sec:discussion}. Finally our paper ends in Section~\ref{sec:conclusions} with suggestions for future work.
    
\section{Background}\label{sec:background}

%\ncn{Appropriate citations needed in this Section - NCN will add them later}

This section presents the core concepts, terminology, and theoretical foundations required to understand the proposed NL framework and its extension to graph-based data.

\subsection{Neurochaos Learning (NL)}\label{subsec:nl}

Neurochaos Learning (NL)~\cite{harikrishnan2021noise} arose out of the observation that chaos and noise are inherent in the brain\cite{NB2021425}. Hence, like causality, NL is also a brain-inspired idea. The brain consists of $\approx 86$ billion neurons that interact with each other to form a complex network~\cite{azevedo2009equal}. These neurons are known to possess fluctuating responses to stimuli. This is partly due to their inherently chaotic nature and also due to noise, which is usually referred to in the literature as ``stochastic resonance (SR)''~\cite{benzi1982stochastic}.

In previous research studies~\cite{harikrishnan2020neurochaos,harikrishnan2021noise,balakrishnan2019chaosnet}, it was shown how SR can be used in NL. Also, the brain is known to be energy efficient and robust to noise, unlike computers. 

%It is well-known that the brain is far better than electronic devices in terms of low computational power, robustness to noise and neural interference. SR in the brain and nervous system could serve as a motivation for the design of robust machine learning architectures and electronic systems. Hence in earlier work it has been shown how SR is used in NL~\cite{harikrishnan2021noise,harikrishnan2020neurochaos,balakrishnan2019chaosnet}. In a nutshell, for SR to occur, the following four elements are required:

The basic learning architecture, \textsf{ChaosNet}, consists of a
layer of chaotic neurons which is the 1D GLS (Generalized Lüroth
Series) map, $C_{GLS}: [0, 1) \rightarrow [0, 1)$,  defined as follows:\\

$C_{GLS}(x) =
\left\{
	\begin{array}{ll}
		\frac{x}{b}  & \mbox{if } 0 \le x < b \\
        \frac{1-x}{1-b} & \mbox{if } b \le x < 1
	\end{array}\label{eq:gls}
\right.
$\\

where $x \in [0, 1)$ and $0 < b < 1$.

Fig.~2 of Ref.~\cite{balakrishnan2019chaosnet} depicts the ChaosNet (NL) architecture. Each GLS neuron starts firing (from an initial neural activity of $q$ units) upon encountering a stimulus (normalized input data sample) and halts when the neural firing matches the stimulus (the topological property of the GLS maps guarantees this for almost every initial $q$). From the neural trace of each GLS neuron, we extract the
following features: {\it firing time, firing rate, energy} and {\it entropy}. These features are aggregated for each class resulting in {\it mean representation vectors} (that are unique for every class). The test sample also undergoes a similar non-linear chaotic transformation resulting in a vector of chaotic features. These learned vectors are then fed either to a simple {\it cosine similarity classifier} (ChaosNet) or to any other {\it machine learning classifier}, or even a neural network, which can also be used (NL-ML hybrid) to yield a classification label.

%. into a classifier such as support vector machine (SVM)~\cite{noble2006support} . 

\begin{comment}
\begin{itemize}
    \item \emph{Firing time}: Time taken for the chaotic trajectory to recognize the stimulus
    \item \emph{Firing rate}: Fraction of time the chaotic trajectory is above the discrimination threshold $b$ so as to recognize the stimulus
    \item \emph{Energy}: For the chaotic neural trace (trajectory) $x(t)$ with firing time $n$, energy is defined as:

    \begin{equation}
        E = \sum_{t=1}^{n}|x(t)|^2.
    \end{equation}
    \item \emph{Entropy}: For the chaotic neural trace (trajectory) $x(t)$, we first compute the binary symbolic sequence $S(t)$ as follows:

        \begin{equation}
            S(t_i) =
            \left\{
            	\begin{array}{ll}
            		0  & x(t_i) < b, \\
                    1 & b \le x(t_i) < 1.
            	\end{array}
            \right.
        \end{equation}
        
        where $i = 1~to~n$ (firing time). We then compute Shannon Entropy of $S(t)$ as follows:

        \begin{equation}
            H(S) = - \sum_{i=1}^{2} p_i log_2(p_i) bits,
        \end{equation}

        where $p_1$ and $p_2$ refer to the probabilities of the symbols 0 and 1 respectively.
\end{itemize}
    
\end{comment}

\subsection{Knowledge Graphs}\label{subsec:kgsem}

A \emph{Knowledge Graph} (KG) is a graph whose nodes correspond to entities and edges correspond to relations among the entities~\cite{ehrlinger2016towards}. Each entity could belong to an \emph{ontology}~\cite{guarino2009ontology}, which defines to which semantic class the entity belongs, and its position in the semantic class. For example, an entity that represents a person could belong to the category of users of a computer system, and the users could themselves be organized in a hierarchy which defines the position of the person in question in the ontology, viz., whether the person in question is a worker or manager. The entity could therefore possess certain attributes (e.g., name, ID, position in the hierarchy, department to which the person belongs) which can then be used to define \emph{features} for the node representing the entity. These features can be grouped together as a \emph{feature vector} representing the node in question. 

Similarly, an edge in the KG represents a relation between the two entities that it links together. Examples of relations could be sibling, colleague, manager, subordinate, etc. This relation can also be attributed just like an entity.

KGs have proved to be highly versatile in knowledge representation and are being used extensively in industry~\cite{velasquez2025neurosymbolic}. This is because the KG representation is problem-independent and can be reused to solve any kind of problem that industry faces.

KGs have been researched quite extensively~\cite{pirro2019building,liu2022multi,douglas2022learned,fionda2020learning}. Several algorithms for aggregating node and edge features, e.g.,~\cite{yang2020nenn}, have also been proposed. 

%\begin{figure*}
%    \centering
    %\includegraphics[width=1.0\linewidth]{chaosnet.jpg}
    %\caption{\textsf{ChaosNet} Architecture }
%    \label{fig:chaosnet}
%\end{figure*}

\subsubsection{Graph-Based Learning and Node Classification}
Numerous real life datasets exist as graphs; nodes designate entities and edges connecting them. The objective of node classification consists of assigning a label to every individual node within the graph, taking advantage of both the attributes associated with each individual node and the underlying structure provided by the graph itself.

GNNs (Graph Neural Networks), a typical method of learning Graphs, have proven successful; however, many problems associated with their use include over-smoothing, expensive run-time costs, and sub-par results when applied to heterophilic graphs.

In this paper, we expand upon NL as applied to Graphs without using Graph Neural Networks (GNN) as the basis for this work. The graph structure will be incorporated by using a process called node feature aggregation, whereby the features of a node’s neighbors will be aggregated using a mean aggregation process. The resulting aggregated values will then be combined with the original node attribute values and fed through a Chaos Based Transformation Pipeline. 

\subsubsection{Homophily, Heterophily, and Class Imbalance}
Graph datasets have a large variation in structure. In homophilic graphs, nodes connect with other nodes in their own class, and in heterophilic graphs, nodes connect with nodes in a different class. Such properties have a large effect on learning performance.

Moreover, class imbalances have been seen in many real-world graph data sources, in which a few classes have very few examples. Class imbalances can result in biased learning schemes that adversely impact models when accuracy is used for evaluation.

Neurochaos learning with a macro-average criteria of assessment provides a powerful tool for analyzing these data sets since it highlights good classes and magnifies variations in features.

\subsection{Chaos-Based Transformation Pipeline}
\begin{figure}[t]
\centering
\includegraphics[width=0.85\textwidth]{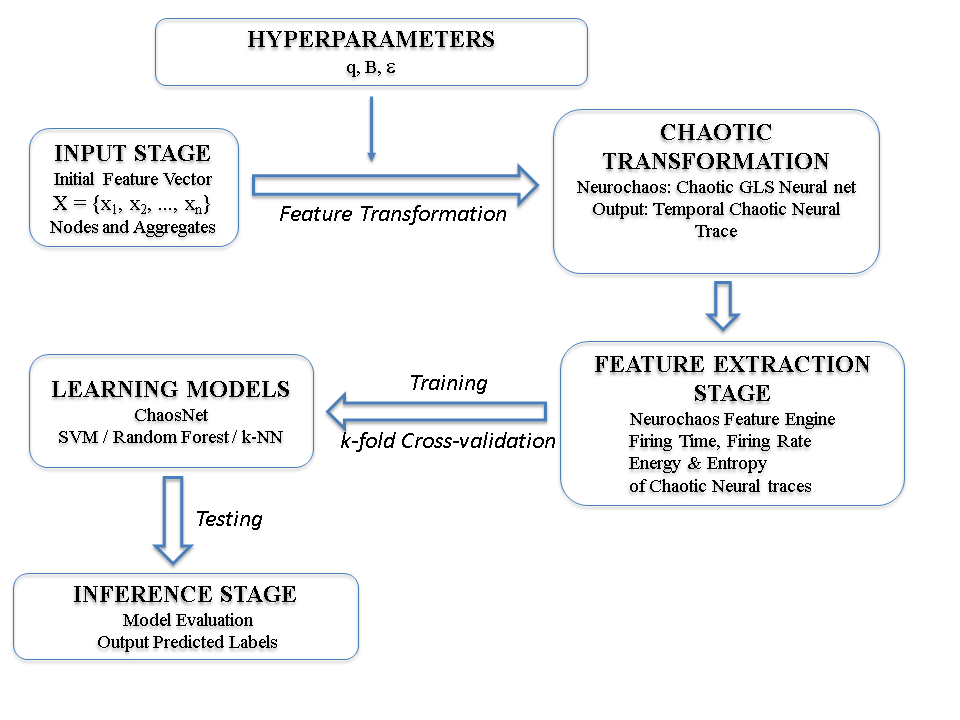}
\caption{Neurochaos-based feature transformation and classification pipeline.}
\label{fig:pipeline}
\end{figure}

The chaos-based transformation network is applied to transform the feature set for nodes to produce an enhanced and discriminative feature space prior to carrying out classification. Rather than applying learning to the feature attributes, every feature undergoes a chaotic dynamical system that generates a non-linear mapping to aid in accentuating the fine but informative details that are usually neglected by traditional machine learning models\cite{AS2023113347}.The overall processing pipeline of the proposed approach is illustrated in Figure.~\ref{fig:pipeline}.

In this pipeline, ``every feature of the nodes in the network serves as an input signal to a GLS-based chaotic neuron.''~\cite{sethi2023neurochaos} When the input signal is presented, the ``chaotic neuron produces a time series response, which is termed a 'neural trace.’ The dynamics of the trace are controlled by three parameters: the {\it initial activity level} of the neuron, the value of the {\it discrimination threshold}, and the {\it noise intensity}.'' As in all chaotic systems, ``small variations in the actual input signal result in distinct variations of the corresponding neural response.''

For each neural trace, four chaos-based features are extracted, namely firing time, firing rate, energy, and entropy~\cite{sethi2023neurochaos}. Each of these captures a different range of the chaotic response, such as how fast it takes for the neuron to react, how frequently it crosses over a threshold, the amplitude of the response, and how irregular the signal appears. These collectively result in a compact-informative feature vector per original node attribute.

The pipeline, on the other hand, processes original node features and aggregated neighbor features generated by the mean aggregation for graph-based data. Combination and ChaosNet classification result from the chaos-based features extracted from such inputs. This allows the model to explore both node-level information and graph structures without necessarily reverting to deep graph neural network models.

%<<WE MUST ADD A BLOCK DIAGRAM OF THE PIPELINE>>
%\nithin{We must add a block diagram of the pipeline.}

%\color{black}
\section{Literature Review}\label{sec:litreview}
Learning from linkable and structured data entails the ability to understand the way variables are related and the flow of information from and to the variables. A common way of describing the relationship using a noticeable model is the probabilistic graphical model. A brief description of Bayesian networks as provided by Heckerman~\cite{heckerman2008tutorial} summarizes the application of Bayesian networks as directed acyclic graphs (DAGs). Variables are denoted by the nodes and the relationship between the variables as described by the edges of the graph. One of the merits of Bayesian networks is that a degree of reasoning about the uncertainty and the application to the cause and effect relationship is possible.

Building on these ideas, Scheines~\cite{scheines1993causal} explains the basic principles of causal discovery and highlights the difference between causal prediction and purely statistical prediction. A key concept discussed is d-separation, which connects the structure of a DAG with conditional independence relationships in probability distributions. Scheines further explains that, under assumptions such as the Causal Markov condition and Faithfulness, it is possible to recover the structure of a causal graph from observed data. These works emphasize how graph structure plays a crucial role in understanding data dependencies.

From the perspective of large-scale structured data on the web, Bizer et al.~\cite{bizer2008linked} introduce the concept of Linked Data, which transforms the traditional web of documents into a web of interconnected data entities. They define four core principles: assigning URIs to entities, using HTTP URIs for access, providing structured RDF information, and linking to other related URIs. These principles enable different datasets, such as DBpedia and GeoNames, to be connected and queried together. This work highlights the importance of graph-structured representations for integrating diverse data sources.

To improve explainability in artificial intelligence systems, Jaimini and Sheth~\cite{jaimini2022causalkg} propose the Causal Knowledge Graph (CausalKG) framework. They point out that traditional knowledge graphs mainly represent simple binary relations, which limits deeper reasoning. By combining Bayesian networks with domain ontologies, CausalKG enables the modeling of direct, indirect, and total causal effects. This allows AI systems to answer ``what-if'' questions and provide explanations that are easier for humans to understand, especially in safety-critical applications.

More recently, Jing et al.~\cite{jing2024h2gnn} focus on the issue of learning over graphs with varied structural properties under the paradigm of Hyper Graph Neural Networks (H2GNN)~\cite{jing2024h}\cite{article_1578830}. The authors perceive GNNs learning poorly over heterophilic graphs where similar nodes in the vicinity take diverse categories. To counter this shortcoming, they propose feature decoupling and adaptive aggregation in order to adequately aggregate information from similar as well as different nodes. The authors show effective results of H2GNN on diverse benchmark datasets.

\section{Research Gap}\label{sec:researchgaps}

Even though the aforementioned works have a solid basis, some gaps exist while implementing these concepts for the learning process of NL on connected and graph data.

First, although Bayesian Networks and Causal Structure Learning algorithms discussed in Heckerman~\cite{heckerman2008tutorial} and Scheines~\cite{scheines1993causal} provide powerful symbolic and probabilistic capabilities, they ignore the potential application of chaotic dynamical systems as computational units. In particular, none of these papers consider the application to graph structured representations, nor learning from dependencies that might be addressed by chaos-induced feature transformation, including Neurochaos learning. The combination of Causal Graph concepts with nonlinear chaos-based feature extraction has not been considered until now\cite{HENRY2025117213}.

Second, Linked Data frameworks proposed by Bizer et al.~\cite{bizer2008linked} and causal extensions such as CausalKG~\cite{jaimini2022causalkg} mainly focus on semantic interoperability, ontology design, and reasoning. However, they do not address how learning models based on chaotic neural dynamics can operate directly on such graph-structured data. Existing NL models are primarily designed for tabular or structured data and lack mechanisms to directly capture graph connectivity and neighborhood information present in linked datasets.

Third, although H2GNN~\cite{jing2024h2gnn} demonstrates strong performance on heterophilic graphs using complex dual-aggregation architectures, it remains unclear whether chaos-based features such as firing time, energy, and entropy can provide robustness to heterophily without relying on deep or computationally heavy GNN architectures. There is currently no work that studies whether a {\it lightweight chaos-based feature transformation} can serve as an effective alternative for node classification on heterophilic graphs.

This serves as a reason to explore the NL method as a structure-aware feature transformation technique on graph-structured data. Through the examination of chaotic feature representation under various properties of a graph such as homophily, heterophily, and class imbalance, the goal of the paper is to fill the existing gap between learning using chaos and graph-structured data without relying on complicated graph neural networks.

\section{Objectives}\label{sec:objectives}

In this paper, the research goal is to explore the application of NL for node classification on graph-structured data. This paper will concentrate on analyzing how chaos-based feature transformation techniques can help discover subtle and minute information present in graph structured data, which normally, through traditional machine learning models, would remain undetected. Through the extension of Neurochaos learning towards graph-structured data, it will try to investigate its applications on being trained on the attributes as well as structure of the nodes.

The initial objective of the proposed research work is applying the chaos feature transformation technique on different graph data sets, citation graphs, and social networks. The application of chaotic dynamics in transforming features of nodes in the graph data sets is anticipated to emphasize the minute differences in the data sets.

The second goal is to investigate Neurochaos learning with the combination of original node features and neighborhood-aggregated features. This can provide insights into how the local graph structure contributes to the classification performance and whether the chaos-based transformations can utilize the neighborhood information effectively, especially in graphs with class imbalance or heterophily.

The ultimate goal is to assess the performance of the Neurochaos learning paradigm with respect to the macro-F1 measure on various linked datasets. By means of thorough hyperparameter optimization, this work intends to investigate learning dynamics influenced by features of graphs, as well as test if a more straightforward approach to machine learning on graphs might be achieved with a graph neural network learning paradigm such as NL.

\section{Methodology}\label{sec:methodology}

In this study, we primarily focused on NL and investigated how it can be effectively applied to linked and graph-structured datasets. Subsequently, we worked with several well-known linked datasets, including {\it Cora, Citeseer, Pubmed, Actor, Cornell, Wisconsin} and {\it Squirrel}~\cite{zhang2024chaotic}. These datasets consist of nodes,edges and node-level features. Since NL operates only on tabular feature matrices and does not directly utilize graph structures we first transformed the graph data into a tabular representation. In this format, each node corresponds to a single data instance and is associated with a feature vector and a class label.

%Initially, we began by executing the basic NL implementation~\cite{harikrishnan2020neurochaos} on small-scale datasets to understand the underlying chaotic feature extraction mechanism and to study the influence of key parameters such as $Q$, $B$, and $\epsilon$ on the generated features. During this phase, we also experimented with a combination of NL and Decision Trees (DT). However, the obtained results were inconsistent across datasets and did not demonstrate reliable performance. As a result, we decided to shift our focus toward graph-based datasets.

To incorporate graph information without relying on Graph Neural Networks 
(GNNs)~\cite{jing2024h2gnn},
we employed a node aggregation strategy. For each node the mean of its neighboring nodes’ feature vectors was computed. Based on this approach we evaluated three different input configurations:
(i) original node features, (ii) aggregated neighbor features and (iii) a combination of original and aggregated features referred to as dual loading. Experimental results indicated that the dual loading approach generally improved performance. However, for certain heterophilic datasets like Actor, feature aggregation resulted in reduced performance.

Before applying NL all feature values were normalized to the range $[0,1]$, because chaotic feature extraction requires normalized inputs. The node identifier column was removed and label consistency was verified. A stratified 80-20 train--test split was used in order to preserve class distribution across both training and testing sets ensuring fair evaluation when class imbalance is present.

ChaosFEX~\cite{sethi2023neurochaos} was then applied to extract chaotic features from the normalized data. Each original feature was transformed into four chaotic features,leading to a significant increase in feature dimensionality. The primary parameters involved in ChaosFEX include $Q$ (initial neural activity), $B$ (discrimination threshold), and noise intensity $\epsilon$. These parameters were tuned individually for each dataset, as a single configuration did not yield optimal results across all graph datasets.

For classification, we utilized ChaosNet~\cite{balakrishnan2019chaosnet} instead of conventional machine learning classifiers. ChaosNet computes the mean feature vector for each class during training and predicts the class of a test sample using a simple {\it cosine similarity classifier}. This approach proved to be particularly effective in scenarios with limited training data and class imbalance.

Hyperparameter tuning was performed using $k$-fold cross-validation ($k=5$), and performance was evaluated using the macro F1-score, as accuracy is not a reliable metric for imbalanced datasets. Experimental observations revealed that datasets with high homophily,such as Cora and Pubmed  benefited significantly from feature aggregation, whereas datasets with high heterophily, such as Squirrel, Cornell, and Wisconsin, showed degraded performance with aggregation.

All experiments were conducted using Python, NumPy, Scikit-learn, and PyTorch Geometric. The numerical calculations and matrix operations needed for chaos-based feature transformation were performed using NumPy and SciPy. Preprocessing node features, organizing experimental results, and reading dataset files were all done with Pandas. Train-test splitting, k-fold cross-validation ($k = 5$), and the computation of evaluation metrics such as macro F1-score were all done using Scikit-learn. Graph datasets were loaded using PyTorch Geometric, and node feature aggregation based on graph structure was carried out. Multiple experimental runs were carried out for each dataset, and the best performing hyperparameters were selected based on validation macro F1-score. Overall, this work involved an in-depth understanding of NL, adapting it for linked data, evaluating multiple feature configurations, and analyzing the impact of graph structure on classification performance.

\subsection{Dataset Acquisition}
\label{subsec:dataset_acquisition}\label{sec:datasets}

The proposed NL-based framework was evaluated on multiple benchmark node classification datasets loaded using the PyTorch Geometric library. These datasets vary in terms of the number of nodes, the dimensionality of node features, the number of classes, and the underlying graph structure. A key distinction among these datasets is whether they exhibit \emph{homophily} or \emph{heterophily}.

To quantify homophily and heterophily, we adopt a standard formulation widely used in the graph learning literature. Let $G = (V, E)$ denote a graph, where $V$ is the set of nodes and $E$ is the set of edges. Each node $v \in V$ is associated with a categorical class label $c(v)$. For each edge $(u, v) \in E$, we define:
\begin{itemize}
    \item $E_{\text{same}}$: the number of edges connecting nodes with the same label, i.e., $c(u) = c(v)$,
    \item $E_{\text{diff}}$: the number of edges connecting nodes with different labels, i.e., $c(u) \neq c(v)$,
    \item $E_{\text{total}}$: the total number of edges in the graph.
\end{itemize}

The homophily ratio is defined as
\[
\text{Homophily} = \frac{E_{\text{same}}}{E_{\text{total}}},
\]
which measures the proportion of edges connecting nodes belonging to the same class. Correspondingly, the heterophily ratio is defined as
\[
\text{Heterophily} = \frac{E_{\text{diff}}}{E_{\text{total}}}
= 1 - \text{Homophily}.
\]

These scores are used in this work to determine whether a dataset should be considered homophilic or heterophilic (and if so, how heterophilic it is), rather than relying solely on qualitative descriptions.

All datasets were loaded directly using the PyTorch Geometric dataset loaders. The library performs basic pre-processing automatically, including the removal of duplicate edges and self-loops, validation of node labels, and verification that node feature matrices do not contain missing values. This ensured consistency across datasets prior to any additional processing steps.

A consolidated quantitative summary of all datasets used in this study is presented in Table~\ref{tab:dataset_summary}. This Table also shows the homophily metrics for each dataset, which shows that Cora, Pubmed and Citeseer are highly homophilic, while the others are highly heterophilic. 

%\ncn{Homophily \& heterophily scores for Citeseer should be shown in Table~\ref{tab:dataset_summary}(we have addressed this).}

\begin{table}[htbp]
\centering
\caption{Quantitative summary of linked datasets used in this study.}
\label{tab:dataset_summary}
\small
\setlength{\tabcolsep}{4pt}
\begin{tabularx}{\linewidth}{lccccccX}
\hline
\textbf{Dataset} & \textbf{Nodes} & \textbf{Features} & \textbf{Edges} & \textbf{Classes} & \textbf{Hom.} & \textbf{Het.} & \textbf{Graph Property} \\
\hline
Cora      & 2,708  & 1,433 & 5,429   & 7 & 0.810 & 0.190 & Highly homophilic \\
Citeseer  & 3,327  & 3,703 & 4,732   & 6 & 0.738    & 0.262   & Moderately homophilic \\
PubMed    & 19,717 & 500   & 44,338  & 3 & 0.802 & 0.198 & Moderately homophilic \\
Actor     & 7,600  & 932   & 30,019  & 5 & 0.219 & 0.781 & Highly heterophilic \\
Cornell   & 183    & 1,703 & 295     & 5 & 0.131 & 0.869 & Highly heterophilic \\
Wisconsin & 251    & 1,703 & 499     & 5 & 0.196 & 0.804 & Highly heterophilic \\
Squirrel  & 5,201  & 2,325 & 217,073 & 5 & 0.224 & 0.776 & Highly heterophilic \\
\hline
\end{tabularx}
\end{table}

To make the graph datasets compatible with the NL framework, they were converted into a CSV-based format using PyTorch Geometric. Following dataset loading, node features and labels were taken out of the graph data object and put into a file, where a single node's feature values and class label are represented by each row. In a similar manner, the edge index provided the graph connectivity data, which was then saved in an \color{blue} $\langle edge~source~node, edge~target~node \rangle$ \color{black} pairs in a CSV file. While allowing for additional processing like node aggregation and chaotic feature extraction, this conversion maintains the original graph structure. To make this conversion clearer, consider a small example from the Cora citation dataset. In Cora, each node represents a research paper, and edges represent citation relationships between papers. Each paper is represented using a bag-of-words feature vector and a class label
corresponding to the research topic. For example, if Paper 12 cites Paper 45, and Paper 45 in turn cites Paper 87, when transformed into CSV format, this creates two files. The nodes.csv file includes one row per paper, with the row including the paper ID, feature
values, and the class label. The edges.csv file includes the citation relationships, with each row in the table including a citation relationship such as (12,45) or (45,87).
This format includes the content information for each paper as well as the relationships between papers in the form of citations, with the data being in a form that can be processed using the NL framework.

\subsection{Initial Exploratory Phase }\label{sec:initial}
In the initial exploratory phase, experiments were conducted on the Cora dataset to analyze the behavior of NL-based models on graph-structured data. These experiments were performed using only node-level features extracted from \texttt{nodes.csv}. We first evaluated a hybrid model that combines a Random Forest classifier with ChaosNet. The hyperparameters $Q$, $B$, and $\varepsilon$ used in this phase were adopted from earlier hybrid model experiments conducted on non-graph datasets and were reused here solely to study the behavior of the hybrid approach on graph datasets.

Using the hybrid ChaosNet model with $Q = 0.49$, $B = 0.75$, and $\varepsilon = 0.0071$, along with a maximum tree depth of $\text{MD} = 18$ and number of estimators $\text{NEST} = 500$, the model achieved a macro training F1-score of $0.8026$ and a corresponding macro testing F1-score of $0.8173$. Although the hybrid model demonstrated reasonable performance, the results were not consistently high and required additional tuning of tree-specific parameters.

Subsequently, ChaosNet was evaluated independently on the same dataset. With $Q = 0.52$, $B = 0.75$, and $\varepsilon = 0.1$, the standalone ChaosNet model achieved a higher macro training F1-score of $0.82827$ and a macro testing F1-score of $0.8437$. Since ChaosNet alone provided superior training performance and more stable results without reliance on an additional classifier, all subsequent experiments were conducted exclusively using ChaosNet. The role and tuning of the parameters $Q$, $B$, and $\varepsilon$ are discussed in detail in Section~\ref{subsec:testing-eval}.

\subsection{Node Feature Aggregation Strategies }\label{subsec:aggstrategies}
In graph datasets, nodes are connected to each other, so the information of neighboring nodes also matters. Using only node features ignores these connections. To include this neighborhood information in a simple way, feature aggregation strategies are used. These strategies help combine node features with information from their neighbors before applying the learning model.

\subsubsection{Mean Node Aggregation }\label{subsubsec:meanagg}
Mean node aggregation was chosen because it is simple and works well with graph data. In this method, the features of a node’s neighboring nodes are averaged and assigned to that node. This helps each node capture local neighborhood information without increasing model complexity. Since ChaosNet works on tabular data, mean aggregation is suitable as it converts graph structure into a simple numerical form that can be easily used for training. 

\subsubsection{H2GNN-Based Aggregation}
H2GNN~\cite{jing2024h2gnn} stands for \textit{Homophily and Heterophily Graph Neural Network}, and it is designed to work well on graphs that contain both homophilic and heterophilic connections. In normal aggregation methods, the features of the neighboring nodes are mixed directly. This method is only effective when the neighboring nodes belong to the same class. However, in heterophilic graphs, the mixing of the features of the neighboring nodes would have a negative impact on the model’s performance, as the neighboring nodes may belong to different classes.

H2GNN has addressed the problem of the mixing of the features of the neighboring nodes by separating the features instead of mixing them. In other words, the algorithm has separated the features of the nodes by using two different feature representations for each node: one for the similar features and one for the dissimilar features. In the process of passing the information of the neighboring nodes to the current node, the algorithm has considered the type of features to be passed, whether they are similar or dissimilar. The similar features have been combined to the homophilic representation, while the dissimilar features have been propagated separately. In other words, the features have been combined in a controlled fashion to allow the homophilic and heterophilic information to be passed without any interference.

\subsection{Input Configuration and Loading Strategies (Ablation Study)}
To understand how different input features affect performance, multiple loading strategies were evaluated. In the first case, only the original node features were used. In the second case, only the aggregated neighbor features were considered. In the third case, both original and aggregated features were combined, which we refer to as \textit{dual loading}.

Experiments on the Cora dataset showed that dual loading achieved the best performance. Using only original features resulted in a macro F1-score of 0.674, while using only aggregated features improved the score to 0.776. When both feature types were combined, the macro F1-score further increased to 0.813. These results indicate that combining node-level features with neighborhood information can improve performance on graph datasets. However, when the same approach was applied to other datasets, performance varied depending on whether the graph structure was homophilic or heterophilic.

\subsection{NL Hyperparameter Framework (K-Fold Based)}

The behavior of the NL model is primarily controlled by three hyperparameters: the initial neural activity $Q$, the discrimination threshold $B$, and the noise parameter $\epsilon$. Since these parameters directly affect the chaotic feature extraction process, it is important to select proper parameters to ensure stable learning. As an initial basis for this investigation, a broad and uniform search range was used to assess the model's behavior with varying parameter values.

For all datasets, hyperparameters were initially tuned within the search space of [0.01, 0.5] with a step size of 0.01. The broad search space was used to explore both low- and high-sensitivity regions of the NL system. The initial tuning of hyperparameters aimed to identify reasonable operating regions of Q, B, and $\epsilon$.

To guarantee proper parameter selection, 5-fold cross-validation (k = 5) was used during tuning. The training data was divided into multiple folds, and different combinations of $Q$, $B$, and $\epsilon$ were evaluated across these folds. One fold was used for validation and four folds were used for training in each iteration. The average macro F1-score across all folds was used to compare configurations. The parameter values that consistently produced higher macro F1-scores were selected for final training and testing. 

\subsection{ChaosNet Testing and Evaluation Protocol}\label{subsec:testing-eval}
After training ChaosNet using the selected hyperparameter values, the model was evaluated on test data. So, to follow the evaluation process of ChaosNet, first the dataset had been split into training data and testing data. The class distribution was preserved in both the sets by a stratified split.

The k-fold cross-validation ($k = 5$) then used the training data to pick the best hyperparameters. The data was divided into k equal parts. During each step, $k-1$ parts were used for training while one part was used for validation. This process was repeated until every partition had served once as a validation set. Averaging the macro F1-score across all the folds yielded the selection of the best hyperparameter values.

After the best hyperparameters were chosen, ChaosNet was again trained using all the training data. Lastly, the separate test data that were not involved in model training and validation are now used for testing. In this way, the evaluation of performance will be fair and unbiased. The same feature representations used during training were applied to the test set to ensure consistency. No additional tuning was performed during testing, and all model parameters were kept fixed.

Model performance was evaluated using the macro F1-score, which provides a balanced assessment by assigning equal importance to all classes. This metric is particularly
suitable for graph datasets having class imbalance. The final macro F1-scores obtained on the test sets were used to compare performance across various datasets and feature vector loading strategies.

\section{Experimental Setup}\label{sec:expts}

All experiments were performed on a workstation running a Linux-based operating system. The operating system had an Intel multi-core processor and 32 GB RAM.
All experiments were performed solely on the CPU, and this is justified by the fact that the proposed NL approach is lightweight and does not require GPU acceleration.

The implementation and experimentation were done using Python 3.10 on Ubuntu operating system. The basic libraries and frameworks that were used in this research include NumPy for computation, Pandas for data handling and preprocessing, and Scikit-learn for evaluation metrics and cross-validation methods. PyTorch and PyTorch Geometric were employed for loading, preprocessing, and handling graph-structured datasets. The ChaosFEX and ChaosNet libraries were used for chaotic feature extraction and classification, respectively. This project is available online\footnote{\url{https://github.com/ayushpatravali/Linked-Data-Classification-using-Neurochaos-Learning.}}

%All experiments were performed within a dedicated Python virtual environment to avoid dependency conflicts and ensure consistency across experimental runs. 

The performance of the ChaosNet model is primarily governed by three hyperparameters: the initial neural activity ($Q$), the discrimination threshold ($B$), and the noise intensity ($\epsilon$). Hyperparameter tuning was performed using a grid search strategy combined with five-fold cross-validation. The tuning procedure was carried out in a staged manner. Initially, a coarse search was conducted to identify promising regions for $Q$, $B$, and $\epsilon$. Subsequently, $Q$ and $B$ were fixed near their optimal values, followed by fine-tuning of $\epsilon$ within the region yielding the best performance. Across all experiments, the number of chaotic iterations during feature extraction was fixed at 5{,}000 to maintain consistency.

\section{Results}\label{sec:results}

This section presents the experimental results obtained using the proposed NL-based framework on multiple graph datasets. The goal of these experiments is not only to report classification performance, but also to study how different aggregation strategies and graph structural properties influence the behavior of ChaosNet.

For all datasets, experimentation began with a standard hyperparameter initialization range of 0.01 to 0.5 with a step size of 0.01 for the NL parameters $Q$, $B$, and $\epsilon$. This range is commonly used in NL-based studies~\cite{harikrishnan2020neurochaos,harikrishnan2021noise} and serves as a strong baseline for observing system behavior across low and high sensitivity regions. The use of a uniform
starting range helped ensure consistency, allowing us to perform a fair comparison. Following the detection of stable operating ranges, the parameter space was further refined to obtain dataset-specific configurations. All evaluations are conducted with respect to macro F1-score, which is appropriate for graph datasets with imbalanced classes.

In Table~\ref{tab:results}, we report the results of experiments conducted with ChaosNet
utilizing mean node aggregation with dual loading. The original features of the node, in addition to the features of the neighboring nodes are aggregated.

\begin{table}[htbp]
\centering
\caption{Mean Node Aggregation Results on All Datasets}
\label{tab:results}
\begin{tabular}{l l c c c c c}
\hline
Dataset & Graph Property & Q & B & $\epsilon$ & Train Macro F1 & Test Macro F1 \\
\hline
Cora & Highly Homophilic & 0.52 & 0.75 & 0.10 & 0.828 & 0.844 \\
PubMed & Homophilic & 0.524 & 0.759 & 0.0987 & 0.78 & 0.76 \\
Citeseer & Homophilic & 0.53 & 0.76 & 0.0985 & 0.705 & 0.732 \\
Cornell & Partial Homo/Hetero & 0.524 & 0.76 & 0.0985 & 0.53 & 0.43 \\
Wisconsin & Partial Homo/Hetero & 0.525 & 0.7625 & 0.0975 & 0.58 & 0.57 \\
Actor & Highly Heterophilic & 0.44 & 0.66 & 0.108 & 0.265 & 0.265 \\
Squirrel & Heterophilic & 0.53 & 0.76 & 0.0985 & 0.38 & 0.34 \\
\hline
\end{tabular}
\end{table}

As can be seen from Table~\ref{tab:results}, the role of aggregation strategy is critical to the performance of ChaosNet. For homophilic graphs such as Cora and PubMed, the mean aggregation strategy works well with ChaosNet, which results in considerable performance improvement. For the partially heterophilic graphs, such as Cornell, the effect of aggregation is still positive but less  compared to the fully heterophilic graph, i.e., Actor.

From the results of the experiments on heterophilic graphs, we can see that, under different conditions, the aggregation strategy used could also have a more critical impact on the performance of ChaosNet. To further investigate the impact of aggregation under different structural conditions, the H2GNN-based aggregation strategy was used to aggregate the neighborhood information. Three typical graphs, i.e., Cora (homophilic), Actor (heterophilic), and Cornell (partially heterophilic/homophilic), were used for the experiment. The results are shown in Table~\ref{tab:h2gnn}.

%The fluctuations of the macro F1-score indicate that the aggregation mechanism used can significantly change the interpretation of the neighborhood information by ChaosNet. This, instead of being a weakness of ChaosNet, points to the significance of the aggregation strategy used.

\begin{table}[htbp]
\centering
\caption{H2GNN Results on Selected Datasets}
\label{tab:h2gnn}
\begin{tabular}{l l c c c c c}
\hline
Dataset & Graph Property & Q & B & $\epsilon$ & Train Macro F1 & Test Macro F1 \\
\hline
Cora & Homophilic & 0.18 & 0.14 & 0.18 & 0.308 & 0.23 \\
Actor & Heterophilic & 0.44 & 0.66 & 0.108 & 0.232 & 0.25 \\
Cornell & Partial Homo/Hetero & 0.13 & 0.15 & 0.20 & 0.263 & 0.232 \\
\hline
\end{tabular}
\end{table}

H2GNN~\cite{jing2024h2gnn} provides a structured method to deal with both homophilic and heterophilic relationships by distinguishing between the information received from similar and dissimilar nodes during aggregation. Nevertheless, the performance of the H2GNN method heavily relies on the characteristics of the dataset being employed. In the case of the Cora dataset, which is highly homophilic, the H2GNN method does not perform well because it is not necessary to separate the homophilic and heterophilic relationships, and it may disrupt the high consistency between the node and neighboring node labels. In the case of the Cornell dataset, which is partially heterophilic, the H2GNN method shows relatively better performance because it can better handle the mixed relationship between nodes. For the Cornell dataset, which contains a mix of homophilic and heterophilic connections, H2GNN shows comparatively better performance, as controlled handling of mixed neighborhood relationships becomes useful. However, Cornell is a small web-page graph with a limited number of nodes and highly imbalanced classes. Although the feature dimensionality is high, the node feature vectors are sparse, meaning that each node activates only a small subset of features. This is because, due to the sparsity and the relatively small size of the graph, neighborhood aggregation may propagate weak or noisy signals, and the performance of even heterophily-aware aggregation methods may not be satisfactory. Similarly, in the Actor dataset, which is heterophilic in nature, the performance may not be satisfactory due to the same reason, i.e., the sparsity and noisy nature of the feature representations may not allow the propagation of meaningful information. These observations suggest that the performance of the H2GNN is heavily dependent on the dataset and that the aggregation methods should be chosen properly according to the structure and features of the data. This also implies that the use of chaotic feature extraction methods and aggregation methods may result in better performance in the future.

In addition, the Cora dataset has been utilized to carry out an in-depth investigation of the hyperparameters of ChaosNet. In particular, the entire space of Q, B, and $\epsilon$ in the standard range (0.01 to 0.5, step 0.01) has been explored and recorded. Figure~\ref{fig:cora} presents the 3D plot of the macro F1-score.

\begin{table}[h]
\centering
\caption{Best ChaosNet Runs on the Cora Dataset}
\label{fig:cora}
\begin{tabular}{c c c c}
\hline
Q & B & $\epsilon$ & TrainMacro F1 \\
\hline
0.52 & 0.75 & 0.10 & 0.8437 \\
0.50 & 0.70 & 0.09 & 0.8130 \\
0.49 & 0.75 & 0.07 & 0.8026 \\
\hline
\end{tabular}
\end{table}

The results indicate a stable high-performance region around Q $\approx$ 0.5, B $\approx$ 0.7-0.75, and $\epsilon$ $\approx$ 0.09-0.10. 

\begin{figure}[htbp]
\centering
\includegraphics[width=0.8\textwidth]{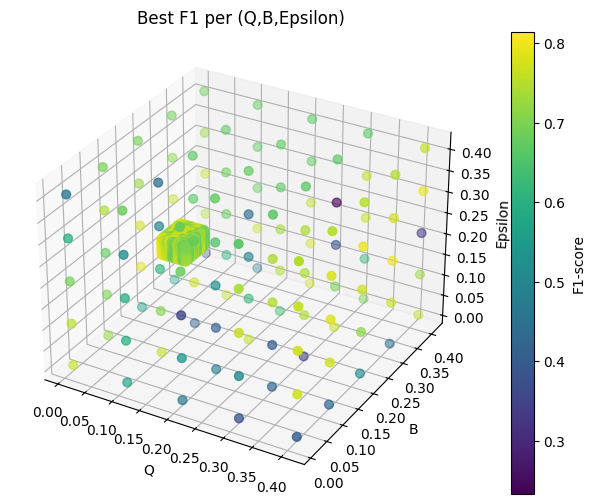}
\caption{3D visualization of macro F1-score across the ChaosNet hyperparameter space on the Cora dataset}
\label{fig:plot}
\end{figure}

Moreover, the 3D plot of the macro F1-score, as depicted in Figure~\ref{fig:plot}, confirms the fact that the performance of the proposed model does not rely on a single point; rather, the model operates within a bounded region of the space of the hyperparameters.
From the experimental results, it is evident that the proposed ChaosNet model is highly sensitive to the representation and aggregation of the input. Significant performance variations are observed simply by changing the aggregation method, without modifying the core learning algorithm. This indicates strong potential for future research in developing aggregation-aware NL frameworks.

The results obtained in this study open multiple research directions, including (a) the exploration of adaptive aggregation strategies, (b) hybrid homophily–heterophily handling, and (c) deeper integration of chaotic feature extraction with graph structural learning.

\section{Discussion}\label{sec:discussion}

In~\cite{harshakokel}, the author describes the six types of neurosymbolic AI systems. Neurosymbolic systems have been proposed as one method to address the well-known limitations of deep learning (as described in~\cite{marcus2022deep}). It uses symbolic models of the domain being analyzed in an effort to incorporate the domain semantics and thereby enhance the accuracy of machine learning. In as much as we have positioned NL as an enhancement of deep learning, our paper actually presents a neurosymbolic extension of NL by incorporating linked data. In our paper, the linked data - which represents the symbolic model of the domain - is aggregated in a manner that is subsequently processed by NL. To that end, NL can be viewed as a subroutine which is invoked within an overall symbolic method; in other words, the Type-2 neurosymbolic system as per the taxonomy of~\cite{harshakokel}. 

However, this is still a limited form of a neurosymbolic system, and the results that we have obtained reflect this fact. In particular, for heterophilic graph datasets, our results need considerable improvement. This calls for further research along two major directions:
\begin{itemize}
    \item Improved node aggregation algorithms that take the actual nature of the heterophily into account, not just the heterophily metric. Here, edge aggregation algorithms that help to integrate edge features, should also be considered to augment the node aggregation algorithms.
    \item Distribution of NL, in particular, specifically designed chaotic maps, across the structure of the knowledge graph, in a manner similar to the chaotic neural oscillators described in~\cite{zhang2024chaotic}.
\end{itemize}

\section{Conclusions and Future Work}\label{sec:conclusions}

In this paper, for the very first time, we have extended Neurochaos Learning (NL) to linked data, i.e., data that is represented in the form of a knowledge graph. We have presented and implemented a node aggregation scheme that aggregate the graph data and feed it into ChaosNet, the NL engine with the simplest architecture (cosine similarity classifier). We have shown the results of our approach for well-known homophilic and heterophilic graph datasets, and we have also discussed our results. 

Future work would involve addressing the research directions raised in Section~\ref{sec:discussion}, viz., improved node and edge aggregation algorithms, and distributing chaotic maps across the knowledge graph dataset.

\bibliographystyle{acm}
\bibliography{refs}
% Add references manually or using BibTeX

\end{document}